\documentclass{article}

\usepackage{microtype}
\usepackage{graphicx}
\usepackage{subcaption}
\usepackage{booktabs} 
\usepackage{caption}
\usepackage{subcaption}
\usepackage{graphicx}
\usepackage{hyperref}

\usepackage[preprint]{icml2026}

\usepackage{amsmath}
\usepackage{amssymb}
\usepackage{mathtools}
\usepackage{amsthm}
\newtheorem{hypothesis}{Hypothesis}

\usepackage[capitalize,noabbrev]{cleveref}

\theoremstyle{plain}
\newtheorem{theorem}{Theorem}[section]
\newtheorem{proposition}[theorem]{Proposition}

\theoremstyle{definition}
\newtheorem{definition}[theorem]{Definition}

\theoremstyle{remark}

\usepackage[textsize=tiny]{todonotes}

\icmltitlerunning{Submission and Formatting Instructions for ICML 2026}

\begin{document}

\twocolumn[
  \icmltitle{Beyond Alignment: Expanding Reasoning Capacity via Manifold-Reshaping Policy Optimization}



  \icmlsetsymbol{equal}{*}

  \begin{icmlauthorlist}
    \icmlauthor{Dayu Wang}{baidu}
    \icmlauthor{Jiaye Yang}{baidu}
    \icmlauthor{Weikang Li}{pku}
    \icmlauthor{Jiahui Liang}{baidu}
    \icmlauthor{Yang Li}{baidu}
\end{icmlauthorlist}

\icmlaffiliation{baidu}{Baidu Inc.}
\icmlaffiliation{pku}{Peking University}

\icmlcorrespondingauthor{Dayu Wang}{2100010872@stu.pku.edu.cn}


  \vskip 0.3in
]



\printAffiliationsAndNotice{}  

\begin{abstract}

Reinforcement Learning with Verifiable Rewards (RLVR) has demonstrated remarkable success in enhancing the reasoning capabilities of Large Language Models (LLMs). However, recent studies question whether RL genuinely expands reasoning capacity or merely aligns existing latent capabilities, arguing that exploration remains confined within the pre-trained model's low-rank bias manifold. In this work, we challenge this accessibility boundary hypothesis by demonstrating that the latent reasoning space can be fundamentally expanded through targeted geometric interventions. We propose \textbf{Manifold-Reshaping Policy Optimization (MRPO)}, a geometric framework designed to fundamentally restructure the inference space of LLMs. MRPO operates in two stages: first, we employ \textit{Spectral Orthogonal Exploration (SOE)} to eject the policy initialization into the null space of the bias manifold; second, we integrate an \textit{Effective Rank} regularization term into the policy optimization objective. This approach incentivizes the discovery and maintenance of high-dimensional reasoning trajectories against the entropy-reducing tendency of standard RL. Empirically, our 4B-parameter method achieves state-of-the-art performance on mathematical tasks, significantly outperforming  larger models (e.g., Qwen3-32B) and expanding the capability boundary beyond standard GRPO. Our code is available at \url{https://anonymous.4open.science/r/MRPO-D57B/}
\end{abstract}

\section{Introduction}

The prevailing paradigm of Reinforcement Learning with Verifiable Rewards (RLVR) has succeeded in aligning Large Language Models (LLMs) with human preferences, but arguably at the cost of lobotomizing their latent reasoning capacity ~\citep{yue2025doesreinforcementlearningreally,zhu2025surprisingeffectivenessnegativereinforcement}. The ``Superficial Alignment Hypothesis'' posits that standard alignment methods---such as PPO  ~\citep{schulman2017proximal} and DPO ~\citep{rafailov2023direct}---act primarily as style transfer mechanisms, eliciting pre-existing capabilities rather than injecting new ones~\citep{zhou2023lima,lin2023unlocking}. However, this alignment imposes a structural ``tax'': by constraining the policy to the low-rank manifold of human-preferred stylistic norms, we inadvertently trap the model within its pre-trained ``knowledge boundary'' ~\citep{lin2024mitigating}. Recent studies confirm that this geometric confinement leads to an ``alignment tax,'' where heavy regularization suppresses the exploration of high-complexity reasoning paths in favor of safe, convergent, but intellectually shallow responses ~\citep{huang2025safety, niu2025mitigating,dou2025improving}. Consequently, standard RL acts as a reasoning ceiling, preventing the discovery of novel solution structures that reside in the model's null space, effectively rendering the most powerful reasoning trajectories geometrically inaccessible.~\citep{zhou2025breaking}

The emergence of pure RL models, exemplified by DeepSeek-R1, challenges this ceiling by removing the alignment constraints entirely, allowing for the emergence of ``Aha moments'' and self-correction behaviors via brute-force exploration ~\citep{guo2025deepseek}. While this approach demonstrates that reasoning capacity is expandable beyond the pre-training mean, it introduces a critical stability gap. Without the geometric tether of alignment, these models suffer from chaotic convergence issues, including language mixing, unreadability, and reward hacking, where the model exploits the reward signal at the expense of logical coherence ~\citep{parmar2025challenges}. The dichotomy is stark: standard RLHF offers stability without expansion (The Trap), while Pure RL offers expansion without steerability (The Gap). There exists no framework that systematically navigates the trade-off between exploring the high-dimensional ``reasoning null space'' and maintaining the structural integrity of the solution manifold.

In this work, we propose \textit{Manifold-Reshaping Policy Optimization} (MRPO), a geometric framework designed to decouple capability discovery from distributional stability. MRPO addresses the reasoning paradox through a two-stage paradigm:
$(i)$ We introduce Spectral Orthogonal Exploration (SOE) as a cold-start training data synthesis pipeline. By projecting policy initializations into the manifold's null space, SOE discovers high-rank reasoning trajectories that are inaccessible to standard greedy search. 
$(ii)$ To prevent reasoning collapse, we integrate an Effective Rank \cite{roy2007effective} spectral reward into the GRPO objective \cite{shao2024deepseekmath}, which explicitly preserves dimensionality in reasoning traces. Empirical results on math benchmarks demonstrate that MRPO not only improves $pass@1$ accuracy but also sustains high performance at large $k$, while maintaining a token cost comparable to standard GRPO, effectively boosting sampling efficiency and expanding the boundary of reasoning capability.

\section{Related Work}

\textbf{Reinforcement Learning with Verifiable Rewards (RLVR).} 
RLVR has emerged as a standard paradigm for reasoning, leveraging deterministic feedback over subjective human preferences~\citep{kaufmann2025surveyreinforcementlearninghuman,guo2025deepseek}. 
However, a central debate persists regarding its mechanism: early studies suggest RLVR merely improves sampling efficiency without expanding the base model's latent capacity~\citep{shao2024deepseekmath,yue2025doesreinforcementlearningreally,zhu2025surprisingeffectivenessnegativereinforcement}, while recent findings argue it can incentivize novel reasoning paths under specific metrics~\citep{chen2025acereasonnemotronadvancingmathcode,wen2025reinforcementlearningverifiablerewards}.
We bridge this gap by offering a geometric perspective: standard RL is bounded by the pre-trained \textit{Bias Manifold}, and genuine capability expansion requires explicitly reshaping this geometry.

\textbf{Dense and Intrinsic Reward Design.} 
To mitigate the sparsity of outcome supervision, prior works introduce dense signals via step-wise process rewards~\citep{yao2026prl,zhang2025linking} or intrinsic motivation based on uncertainty~\citep{zhang2025count} and diversity~\citep{gao2025navigate,yao2025diversity}.
Unlike these approaches, which rely on auxiliary reward models, statistical counts, or confidence scores, our work operates directly on the \textit{latent topology}. 
By optimizing Effective Rank, MRPO provides a physics-grounded intrinsic signal that prevents spectral collapse and maintains high-dimensional information flow without the computational overhead of external value networks.

\textbf{Synthetic Data and Model Collapse.} 
Iterative self-improvement on synthetic data (e.g., STaR, ReST) is foundational for reasoning alignment~\citep{zelikman2022star,gulcehre2023reinforced}, recently popularized by DeepSeek-R1's cold-start specifically~\citep{guo2025deepseek}. 
However, naive self-training risks "model collapse" by over-fitting to the model's own low-entropy distribution~\citep{shumailov2024ai,shafayat2025largereasoningmodelsselftrain}.
We advance this paradigm via \textit{Spectral Orthogonal Exploration (SOE)}. 
Unlike methods that filter for correctness within the existing distribution~\citep{goldie2025syntheticdatageneration}, SOE explicitly injects data from the model's epistemic \textit{Null Space}, mathematically guaranteeing the discovery of trajectories orthogonal to established biases.
\section{Preliminaries: Formalizing the Trap}
\label{sec:preliminaries}

To understand why standard alignment techniques act as a ceiling for reasoning capabilities, we must move beyond behavioral observations and analyze the geometry of the model's latent space. In this section, we provide the mathematical formalization of this problem by defining the \textit{Bias Manifold} and the phenomenon of \textit{Reasoning Collapse}.

\subsection{The Geometry of Latent Representations}
\label{sec:effrank}
Consider an autoregressive language model $\pi_\theta$ generating a reasoning chain $y = (y_1, \dots, y_T)$. Let $h_t \in \mathbb{R}^d$ denote the hidden state vector at the final layer for token $t$. We define the reasoning trajectory matrix $H \in \mathbb{R}^{T \times d}$ as the vertical stacking of these hidden states: $H =^\top$.

To quantify the information content of this trajectory, we utilize the Effective Rank (also known as Stable Rank), a continuous measure of matrix dimensionality derived from the Shannon entropy of its singular values.

\begin{definition}
    Let $\Sigma = \frac{1}{T} (H - \mu)^\top (H - \mu)$ be the covariance matrix of the centered hidden states. Let $\lambda_1 \ge \dots \ge \lambda_d \ge 0$ be the eigenvalues of $\Sigma$. The normalized spectral distribution is given by $p_i = \frac{\lambda_i}{\sum_{j=1}^d \lambda_j}$. The Effective Rank is defined as the exponential of the spectral entropy:
    \begin{equation}
        \text{erank}(H) = \exp\left( - \sum_{i=1}^d p_i \ln p_i \right)
    \end{equation}
\end{definition}

Intuitively, $\text{erank}(H)$ measures the geometric volume of the semantic space explored by the model. A low effective rank implies the trajectory is confined to a low-dimensional subspace, often indicative of simple or repetitive patterns. Crucially, we treat high effective rank as a \textit{necessary condition} for complex reasoning. Therefore, we apply this metric as a regularization term specifically targeting high-quality, \textbf{correct} trajectories, where it serves to distinguish deeper reasoning processes from valid but shallow heuristic solutions.

\subsection{Decoupling Geometry from Uncertainty}
\label{sec:decoupling}

In prior research, token entropy has been widely utilized as a standard proxy for model uncertainty, often serving as the primary metric for filtering high-quality reasoning traces.~\cite{kuhn2023semantic} However, we posit that \textbf{Effective Rank} captures a distinct geometric dimension of reasoning capacity that is independent of probabilistic uncertainty. To validate this hypothesis, we analyze the relationship between geometric complexity, uncertainty, and solution correctness.

\begin{figure}[h]
    \centering
    \includegraphics[width=0.85\columnwidth]{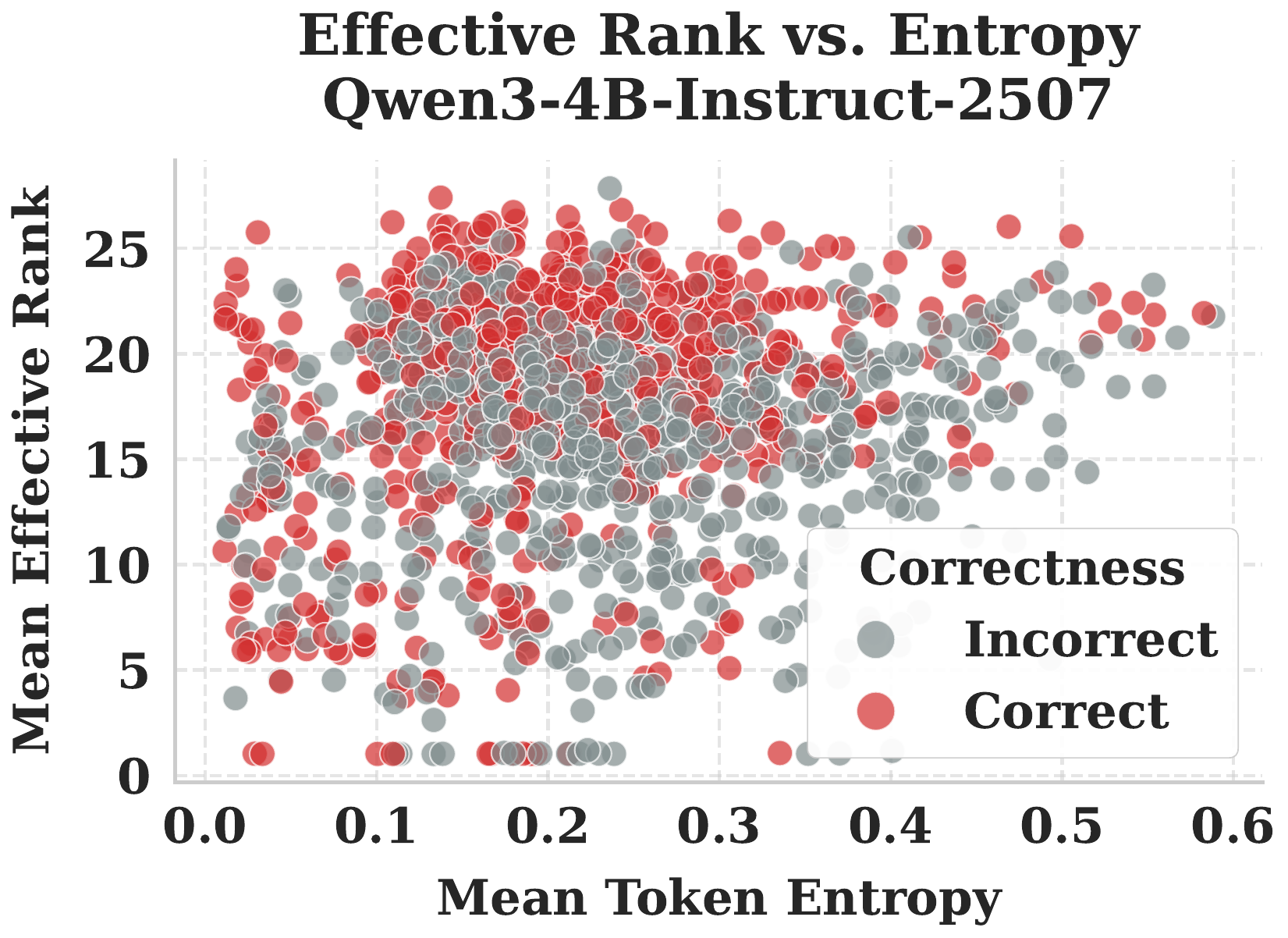}
    \caption{\textbf{Geometric Decoupling of Reasoning Capacity.}}
    \label{fig:rank_entropy}
\end{figure}

Figure~\ref{fig:rank_entropy} visualizes this relationship. We observe that in the high-rank region, the distribution of correct solutions becomes more concentrated than in the low-entropy region.

To rigorously quantify the independent information provided by Effective Rank, we fit a logistic regression model to predict the probability of correctness based on both Effective Rank and Entropy:
\begin{equation}
    P(y \text{ is correct}) = \sigma(\beta_0 + \beta_r \cdot \text{EffRank} + \beta_e \cdot \text{Entropy})
\end{equation}
where $\sigma(\cdot)$ is the sigmoid function. 

\begin{table}[h]
    \caption{\textbf{Statistical Significance Analysis.} Logistic regression results showing the independent contribution of Rank vs. Entropy to correctness.}
    \label{tab:rank_entropy_correlation}
    \centering
    \begin{sc}
    \begin{small}
    \resizebox{\columnwidth}{!}{
        \begin{tabular}{lccccc}
        \toprule
        {Model} & \multicolumn{2}{c}{Rank (Geometry)} & & \multicolumn{2}{c}{Entropy (Uncertainty)} \\
        \cmidrule{2-3} \cmidrule{5-6}
        & Coef ($\beta_r$) & P-value & & Coef ($\beta_e$) & P-value \\
        \midrule
        Qwen3-4B-Inst & \textbf{+0.56} & \textbf{5.4e-06} & & +0.08 & 0.55  \\
        Qwen3-32B     & \textbf{+0.50} & \textbf{0.001}   & & -0.07 & 0.66  \\
        Qwen3-14B     & \textbf{+0.29} & \textbf{0.029}   & & -0.32 & 0.023 \phantom{} \\ 
        \bottomrule
        \end{tabular}
    }
    \end{small}
    \end{sc}
\end{table}

The results in Table~\ref{tab:rank_entropy_correlation} confirm our hypothesis. When controlling for both variables, Effective Rank maintains robust statistical significance ($\beta_r = 0.56, p \ll 10^{-5}$ for Qwen3-4B), whereas entropy loses its predictive power ($\beta_e \approx 0, p > 0.05$). This demonstrates that reasoning capacity is fundamentally a \textit{geometric} property---distinct from, and irreducible to, the model's probabilistic confidence.

\subsection{The Bias Manifold Hypothesis}
\label{sec:bias_manifold}

\begin{figure*}[t] 
    \centering
    \includegraphics[width=0.8\textwidth]{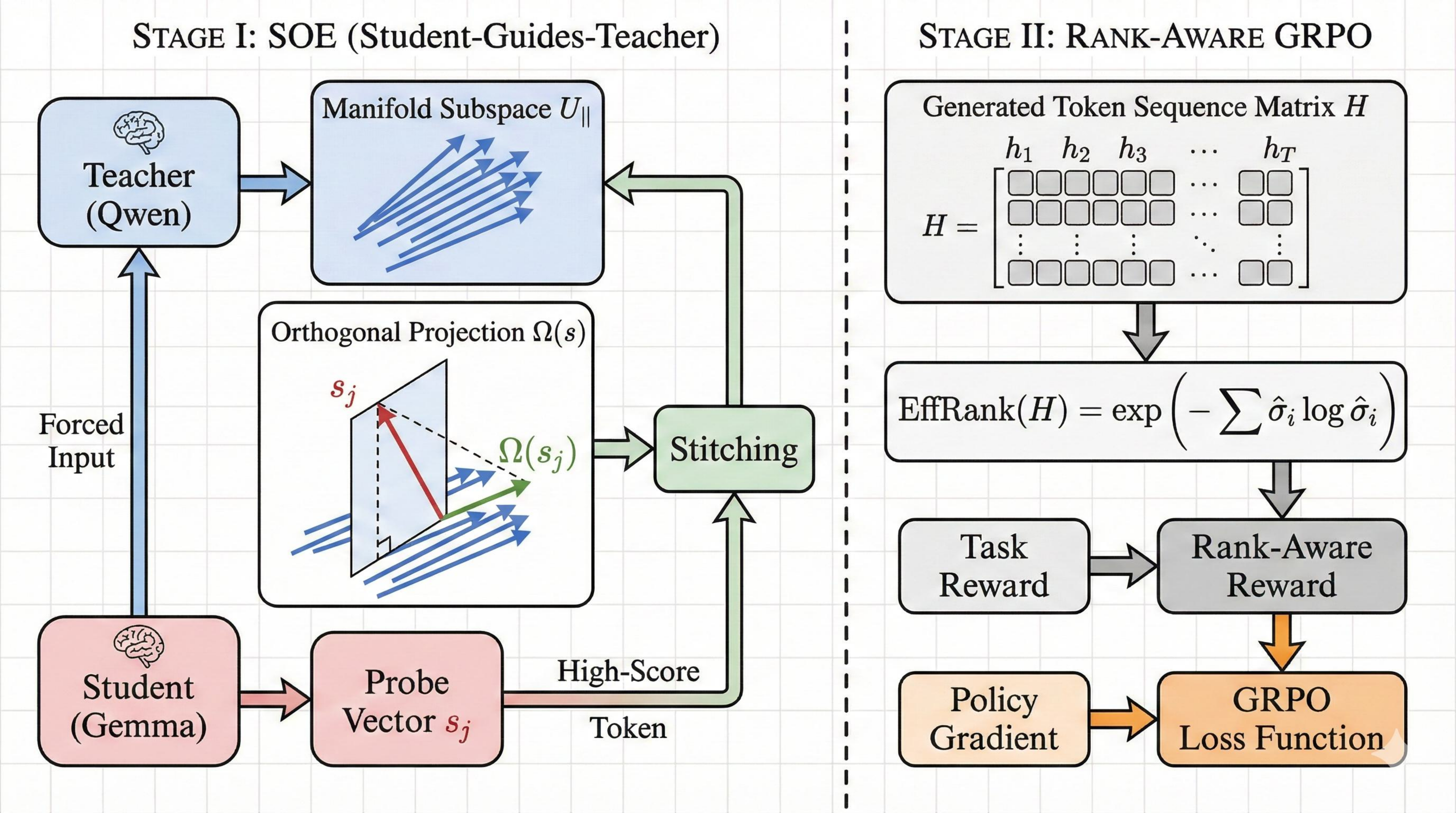} 
    \caption{Geometric Interpretation of \textbf{Our Method's Framework.}}
    \label{fig:reasoning_collapse}    
\end{figure*}

We posit that pre-training and standard SFT compress the vast majority of the model's probability mass into a low-dimensional region dominated by common linguistic patterns and heuristic shortcuts. While the global geometry of LLMs are inherently non-linear, we observe that biased trajectories typically degenerate into a \textbf{Local Linear Subspace}. We term this the \textbf{Bias Manifold}.

\begin{definition}[Local Bias Manifold]
    Consider a localized reasoning trajectory $H \in \mathbb{R}^{T \times d}$. The Bias Manifold $\mathcal{M}_{bias}$ is defined as the subspace spanned by the top-$k$ principal components of $H$, capturing the dominant directions of variation. Formally, we say $H$ is confined to the bias manifold if:
    \begin{equation}
        \frac{\sum_{i=1}^k \lambda_i}{\sum_{j=1}^d \lambda_j} \ge 1 - \epsilon
    \end{equation}
    where $k \ll d$ represents the effective degrees of freedom of the "shortcut" behavior.
\end{definition}

This definition aligns with the ``Superficial Alignment Hypothesis'' \citep{zhou2023lima}: alignment formats the model to stay strictly within $\mathcal{M}_{bias}$. While this confinement ensures high fluency and safety (properties that reside in the dominant eigenspace), it structurally inhibits the exploration of the \textit{null space}---the orthogonal directions where novel reasoning paths and ``Aha moments'' are often found.

\subsection{Reasoning Collapse and Geometric Barriers}
\label{sec:reasoning_collapse}

Standard RL faces a critical \textit{geometric barrier}: complex solutions often require traversing the \textbf{Null Space} $\mathcal{N} = \mathcal{M}_{bias}^\perp$, a region orthogonal to common heuristics. However, the optimization process itself inhibits this exploration due to \textbf{Spectrum Contraction}:

\begin{proposition}[Confidence-Induced Rank Collapse]
    As the policy $\pi_\theta$ converges, increasing logit scales reduce the effective sampling temperature. This acts as a rank-reducing operator, causing the effective rank of generated trajectories to contract:
    \begin{equation}
        \mathbb{E}_{y \sim \pi_\theta}[\text{erank}(H(y))] \to k \ll d
    \end{equation}
    trapping the model in a low-dimensional bias manifold.
\end{proposition}

This formalizes ``Reasoning Collapse,'' leading to \textbf{The Trap}:

\begin{hypothesis}[The Geometric Barrier]
    For a policy confined to $\mathcal{M}_{bias}$, the probability of sampling a trajectory with significant projection onto $\mathcal{N}$ decays exponentially:
    \begin{equation}
        P(\|\text{Proj}_{\mathcal{N}}(H(y))\| > \delta \mid \pi_{init} \in \mathcal{M}_{bias}) < \epsilon
    \end{equation}
    rendering the gradient signal from the Null Space effectively sparse.
\end{hypothesis}

To overcome this inefficiency, we must mechanically ``eject'' the model from $\mathcal{M}_{bias}$ (breaking the ceiling) and subsequently sustain its spectral dimensionality (securing the floor).

\section{Methodology: Manifold-Reshaping Policy Optimization (MRPO)}
\label{sec:methodology}

Based on the geometric theories established in Section 3, we propose \textbf{Manifold-Reshaping Policy Optimization (MRPO)}. Unlike traditional RLHF, which primarily optimizes for reward maximization within a fixed manifold, MRPO actively manages the geometric structure of the policy's latent space. The framework consists of two strictly decoupled stages: \textbf{Stage I: Geometric Ejection} (via SOE Cold-Start) to break the reasoning ceiling, and \textbf{Stage II: Rank-Aware Optimization} (via GRPO) to secure the reasoning floor.

\subsection{Stage I: Geometric Ejection via Spectral Orthogonal Exploration (SOE)}

The objective of Stage I is to construct a high-quality Cold-Start dataset that initializes the policy in a high-rank solution space. We generate 10,000 high-rank reasoning trajectories using the \textit{Spectral Orthogonal Exploration} (SOE) framework proposed by \cite{wang2026studentguidesteacherweaktostrong}. To maximize geometric diversity, SOE employs a Student-Guides-Teacher paradigm where a weaker model acts as an orthogonal probe to help the strong model escape its pre-trained Bias Manifold. Specifically, we utilize Gemma-3-4B-IT ~\cite{gemma_2025}(Student) to assist Qwen3-4B-Instruct-2507 ~\cite{qwen3technicalreport}(Teacher), ensuring that their bias manifolds remain non-overlapping and facilitate effective exploration.

The synthesis mechanism relies on the \textbf{Orthogonal Latent Stitching (OLS)} process. For a given incorrect reasoning trace $x$ from the teacher, we use the first t tokens $x_{<t}$. Use the $x_{<t}$ as context, the Teacher first generates $N=8$ look-ahead samples. We collect their hidden states to form a centered matrix $H$, compute the Gram matrix $G = HH^\top$, and use a Micro-SVD algorithm to extract the principal components $U_{||}$ spanning the Teacher's current local bias manifold. Subsequently, the Student generates $M=8$ candidate reasoning fragments (probes) $s_j$, which are mapped into the Teacher's latent space to obtain vectors $z_j$. We calculate the orthogonal projection residual relative to the Teacher's bias manifold using the formula:
\begin{equation}
    \Omega(s_j) = \frac{\| (I - U_{||}U_{||}^\top)(z_j - \mu) \|}{\| z_j - \mu \| + \epsilon}
\end{equation}
where $\Omega(s_j) \in [0, 1]$ and a value of $\Omega \approx 1$ indicates the probe lies in the Teacher's \textit{Null Space}. We select the probe $s^*$ maximizing $\Omega(s)$ and forcibly stitch its token sequence into the Teacher's context, geometrically ejecting the Teacher from its local optimum to a new coordinate in the Null Space, then we use the new trace as context for teacher to sample.

We iteratively apply this mechanism on the AIME (pre-2023), AMC (pre-2023)~\cite{li2024numinamath}, and MATH training~\cite{lightman2023lets} sets with a high sampling budget of $n=16$ per problem to ensure the discovery of valid high-dimensional paths. Traces are strictly verified using a symbolic math evaluator, and only traces that are both correct and possess high orthogonality scores are retained. Finally, we select 10,000 of these high-quality trajectories to fine-tune the Teacher model for 1 epoch. The Supervised Fine-Tuning (SFT) acts as a geometric rotation, shifting the policy initialization $\pi_{init}$ from $\mathcal{M}_{bias}$ to $\mathcal{N}$ prior to the reinforcement learning stage.

\subsection{Stage II: Rank-Aware Group Relative Policy Optimization}

While Cold-Start provides access to the Null Space, standard RL tends to optimize for minimal-entropy paths, leading to re-collapse. Stage II counters this by integrating a spectral regularization term into the GRPO objective.

\subsubsection{Effective Rank Reward}
We introduce a differentiable metric to quantify the richness of a reasoning chain. We utilize the \textbf{Effective Rank (EffRank)} mentioned in ~\ref{sec:effrank},

To penalize local collapse, we compute EffRank using a sliding window ($w=64$) and use the \textit{minimum} rank over the trajectory as the reward signal. The total reward function is defined as:
\begin{equation}
    R_{total}(y) = \mathbb{I}(y \text{ is correct}) \cdot (1.0 + \alpha \cdot \text{NormRank}(y))
\end{equation}
where $\alpha=0.5$ and $\text{NormRank}$ scales the rank to $[0, 1]$. This incentivizes the model to solve problems using high-dimensional (complex) reasoning paths rather than low-rank shortcuts.

\subsubsection{Rank-Aware GRPO Objective}

To counter the structural bias in the policy space, we optimize the policy using Group Relative Policy Optimization with a rank-augmented reward. For a set of sampled outputs $\{y_i\}_{i=1}^G$ for a given query $q$, we first define the group-relative advantage $\hat{A}_i$ as:
\begin{equation}
    \hat{A}_i = \frac{R_{total}(y_i) - \bar{R}}{\sigma_R}
\end{equation}
where $R_{total}$ incorporates both objective correctness and our proposed geometric richness metric. The resulting Rank-Aware GRPO objective is formulated as:
\begin{equation}
\begin{aligned}
    \mathcal{L}_{GRPO}(\theta) = -\mathbb{E}_{q \sim P(Q), \{y_i\} \sim \pi_{old}} &  \frac{1}{G} \sum_{i=1}^G \hat{A}_i \nabla_\theta \log \pi_\theta(y_i|q)\phantom{\sum}\
\end{aligned}
\end{equation}
The advantage term is now driven by both correctness and geometric richness. This introduces a \textbf{Geometric Prior} that explicitly counteracts the Spectrum Contraction Theorem, forcing gradient flow towards high-rank regions of the parameter space.

\section{Experiments}
\begin{table*}[t]
\centering
\caption{\textbf{Uniform Superiority.}}
\label{tab:main_results}
\label{tab:main_results}
\begin{tabular}{l|cccccc}
\toprule
\textbf{Model} & \textbf{AIME 24} & \textbf{AIME 25} & \textbf{MATH-500} & \textbf{OlympiadBench} & \textbf{Omni-Hard} & \textbf{Mean} \\
\midrule
Gemma-3-4B-IT & 13.3\% & 6.7\% & 71.2\% & 27.7\% & 4.2\% & 24.6\% \\
Qwen3-4B & 40.0\% & 16.7\% & 68.2\% & 30.0\% & 3.0\% & 31.6\% \\
Qwen3-4B-Instruct-2507 & \underline{46.7\%} & 33.3\% & 84.8\% & \underline{42.6\%} & 14.9\% & 44.5\% \\
Qwen3-4B-Instruct-2507 + GRPO & \underline{46.7\%} & \underline{36.7\%} & \underline{87.6\%} & 42.1\% & \underline{16.8\%} & \underline{46.0\%} \\
\midrule
\textit{Reference Models (Higher Scale)} & & & & & & \\
Qwen3-8B & 26.7\% & 20.0\% & 65.2\% & 28.6\% & 4.4\% & 29.0\% \\
Qwen3-14B & 40.0\% & 16.7\% & 81.6\% & 35.7\% & 8.7\% & 36.5\% \\
Qwen3-32B & 33.3\% & 30.0\% & 79.8\% & 35.3\% & 10.8\% & 37.8\% \\
\midrule
\textbf{MRPO (Ours)} & \textbf{56.7\%} & \textbf{43.3\%} & \textbf{88.8\%} & \textbf{43.0\%} & \textbf{17.4\%} & \textbf{49.8\%} \\
\bottomrule
\end{tabular}

\vspace{0.5cm} 

\includegraphics[width=0.8\linewidth]{"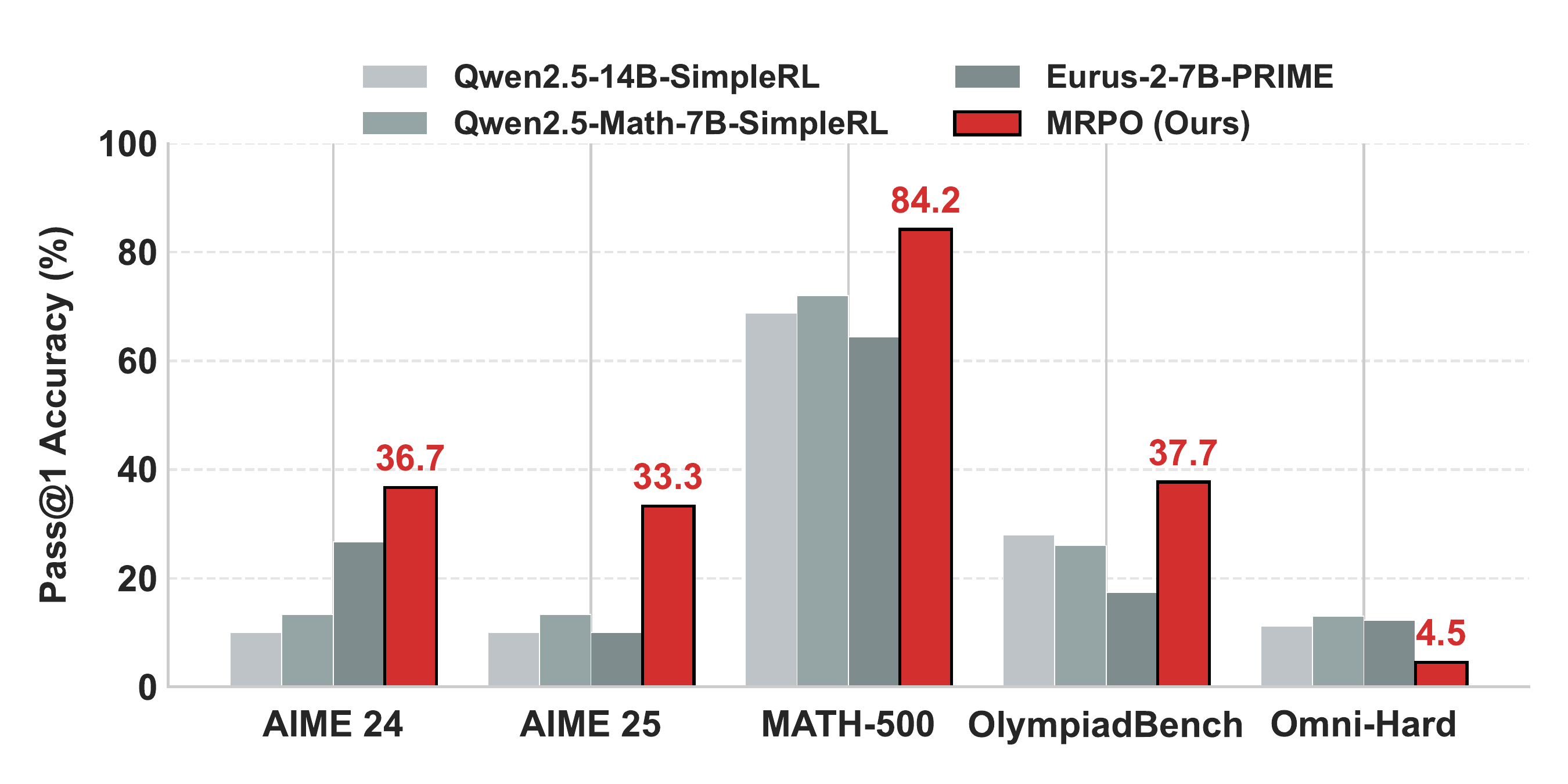"} 
\captionof{figure}{Visualization of performance comparison. MRPO demonstrates superior accuracy on AIME and MATH benchmarks compared to baselines.}
\label{fig:performance_chart}

\end{table*}

We evaluate MRPO on complex mathematical reasoning benchmarks to validate its ability to break the ``Capacity Ceiling'' while securing the ``Stability Floor'' (Rank consistency).

\subsection{Experimental Setup}
\textbf{Benchmarks}  We evaluated the model accuracy on AIME 2024, AIME 2025, MATH-500, OlympiadBench~\cite{he2024olympiadbench}, and Omni-Math~\cite{gao2024omnimathuniversalolympiadlevel}. For Omni-Math, due to its large scale, we evaluate the subset of problems with difficulty level higher than 7 and refer it as Omni-Math (Hard).

\textbf{Baselines.} To rigorously evaluate MRPO, we compare our method against four distinct categories of baselines. First, we examine the \textbf{Base Model}  used for initialization and \textbf{SOE}, \textbf{Qwen3-4B-Instruct-2507} and \textbf{Gemma-3-4B-IT}. Second, we employ a \textbf{Standard GRPO} baseline (Qwen3-4B + GRPO) to control for gains solely attributable to basic reinforcement learning. Third, to test geometric efficiency against parameter scaling, we compare against \textbf{Larger Reference Models} from the same family, including \textbf{Qwen3-8B}, \textbf{Qwen3-14B}, and \textbf{Qwen3-32B}. Finally, we benchmark against state-of-the-art open-source reasoning models, specifically \textbf{SimpleRL} (Qwen2.5-14B / Math-7B)~\cite{zeng2025simplerl} and \textbf{Eurus-2-7B-PRIME}~\cite{cui2025process}. Evaluations for these first three categories were conducted utilizing identical prompts and greedy decoding with a maximum context length of 8192. Finally, we benchmark against state-of-the-art open-source reasoning models. Due to the architectural constraints of \textbf{Qwen2.5-Math-7B-SimpleRL} and \textbf{Eurus-2-7B-PRIME}, which support a limited context window, the comparative evaluations for these reinforcement learning baselines were set to identical prompts and greedy decoding, but with a maximum context length restricted to 4096.

\subsection{Main Results}

\begin{table*}[h]
\caption{Ablation Study. }
\centering
\resizebox{\textwidth}{!}{
\begin{tabular}{l|cccccc}
\toprule
\textbf{Configuration} & \textbf{AIME 24} & \textbf{AIME 25} & \textbf{MATH-500} & \textbf{Olympiad} & \textbf{Omni-Hard} & \textbf{Mean} \\
\midrule
Base (Qwen3-4B-Inst) & 46.7\% & 33.3\% & 84.8\% & 42.6\% & 14.9\% & 44.5\% \\
\midrule
Pure GRPO & 46.7\% & 36.7\% & 87.6\% & 42.1\% & 16.8\% & 46.0\% \\
Rank Reward GRPO (No SOE) & 50.0\% & 30.0\% & \textbf{89.6\%} & 42.3\% & 15.9\% & 45.6\% \\
Cold Start (SOE Only) & \underline{53.3\%} & 36.7\% & 86.2\% & 41.4\% & 13.0\% & 46.1\% \\
Cold Start + GRPO (No Rank) & \underline{53.3\%} & \textbf{43.3\%} & 87.8\% & \underline{42.9\%} & \textbf{17.6\%} & \underline{49.0\%} \\
\midrule
\textbf{MRPO (Ours)} & \textbf{56.7\%} & \textbf{43.3\%} & \underline{88.8\%} & \textbf{43.0\%} & \underline{17.4\%} & \textbf{49.8\%} \\
\bottomrule
\end{tabular}
}
\label{tab:ablation}
\end{table*}

Table \ref{tab:main_results} and Figure \ref{fig:performance_chart} present a comprehensive evaluation of MRPO against internal baselines, scaling reference models, and state-of-the-art (SOTA) reinforcement learning models. The results highlight three key findings regarding geometric efficiency and reasoning capacity.

\textbf{Efficiency Over Scale.} 
As detailed in Table \ref{tab:main_results}, MRPO significantly outperforms the scaling laws that typically govern language model performance. Our 4B-parameter model achieves an impressive \textbf{56.7\%} on AIME 2024, surpassing the significantly larger \textbf{Qwen3-32B} (33.3\%) by a margin of 23.4\%. This empirical evidence challenges the assumption that reasoning capacity is strictly bound by parameter count, validating our hypothesis that "latent capacity" can be unlocked through geometric reshaping rather than mere scale. MRPO secures the highest mean score of \textbf{49.8\%} across all benchmarks, outperforming the standard GRPO baseline (46.0\%) and demonstrating that the gains stem from our specific manifold optimization rather than the generic RL algorithm.

\textbf{Superiority Over SOTA RL Models.} 
Figure \ref{fig:performance_chart} further contrasts MRPO with leading open-source reasoning models, including SimpleRL (7B/14B) and Eurus-2-7B-PRIME. Despite being smaller in size, MRPO establishes a clear dominance on the first four datasets. Notably, on the rigorous \textbf{MATH-500} benchmark, MRPO achieves \textbf{84.2\%} (as visualized in the figure comparisons), significantly outpacing the 14B-parameter baselines. This confirms that explicitly optimizing the effective rank of the policy yields a more robust reasoner than implicit process rewards or rule-based reinforcement alone.

\textbf{Analysis of Context Sensitivity and Truncation.}
A specific anomaly is observed on the \textbf{Omni-Hard} benchmark, where MRPO's lead narrows or dips compared to specific SOTA baselines. This can be attributed to the architectural divergence in training protocols. MRPO is trained with a long-context window of 8192 tokens to encourage deep, high-rank reasoning trajectories. However, when benchmarked against SOTA RL models (which are often optimized for shorter, constrained contexts of 4096 tokens), this introduces a comparative disadvantage on extremely lengthy problems. The "long-chain" reasoning strategy that MRPO develops—while beneficial for depth—renders it susceptible to generation truncation under strict evaluation limits. Consequently, a portion of correct but lengthy derivations on Omni-Hard were cut off, artificially lowering the accuracy metric despite the model's underlying capability.

\subsection{Ablation Study}

To rigorously validate the individual contributions of Spectral Orthogonal Exploration (SOE) and Rank-Aware Regularization, we conducted an ablation study across five mathematical benchmarks. All evaluations reported in Table \ref{tab:ablation} were performed under identical conditions, utilizing greedy decoding with a maximum context length of 8192 tokens to ensure fair comparability.

\textbf{The Limitations of Standard RL.} As shown in the results, applying \textbf{Pure GRPO} directly to the base model yields only marginal improvements, raising the mean accuracy from 44.5\% to 46.0\%. This suggests that without structural intervention, standard reinforcement learning is confined by the pre-trained model's bias manifold, struggling to discover novel reasoning paths.

\textbf{The Necessity of Geometric Ejection (Cold Start).} The \textbf{Cold Start (SOE Only)} configuration achieves a mean accuracy of 46.1\%, effectively matching the performance of Pure GRPO solely through supervised fine-tuning on high-rank trajectories. This confirms that "ejecting" the model into the null space acts as a critical initialization step, establishing a higher capability baseline before optimization.

\textbf{Synergy of Components.} The most significant gains arise when combining these stages. \textbf{Cold Start + GRPO (No Rank)} surges to 49.0\%, demonstrating that once the "accessibility boundary" is broken via SOE, reinforcement learning can effectively exploit the new solution space. Finally, our full method, \textbf{MRPO}, achieves the highest mean performance of \textbf{49.8\%}, with notable gains on AIME 24 (56.7\%). By integrating the Rank-Aware Reward, MRPO prevents the reasoning collapse often seen in standard RL, proving that both the "ejection" (SOE) and the "stability" (Rank Reward) mechanisms are indispensable for maximizing reasoning capacity.

\subsection{Reward Effectiveness Analysis}
\begin{figure}[h]
    \centering
    \includegraphics[width=1\linewidth]{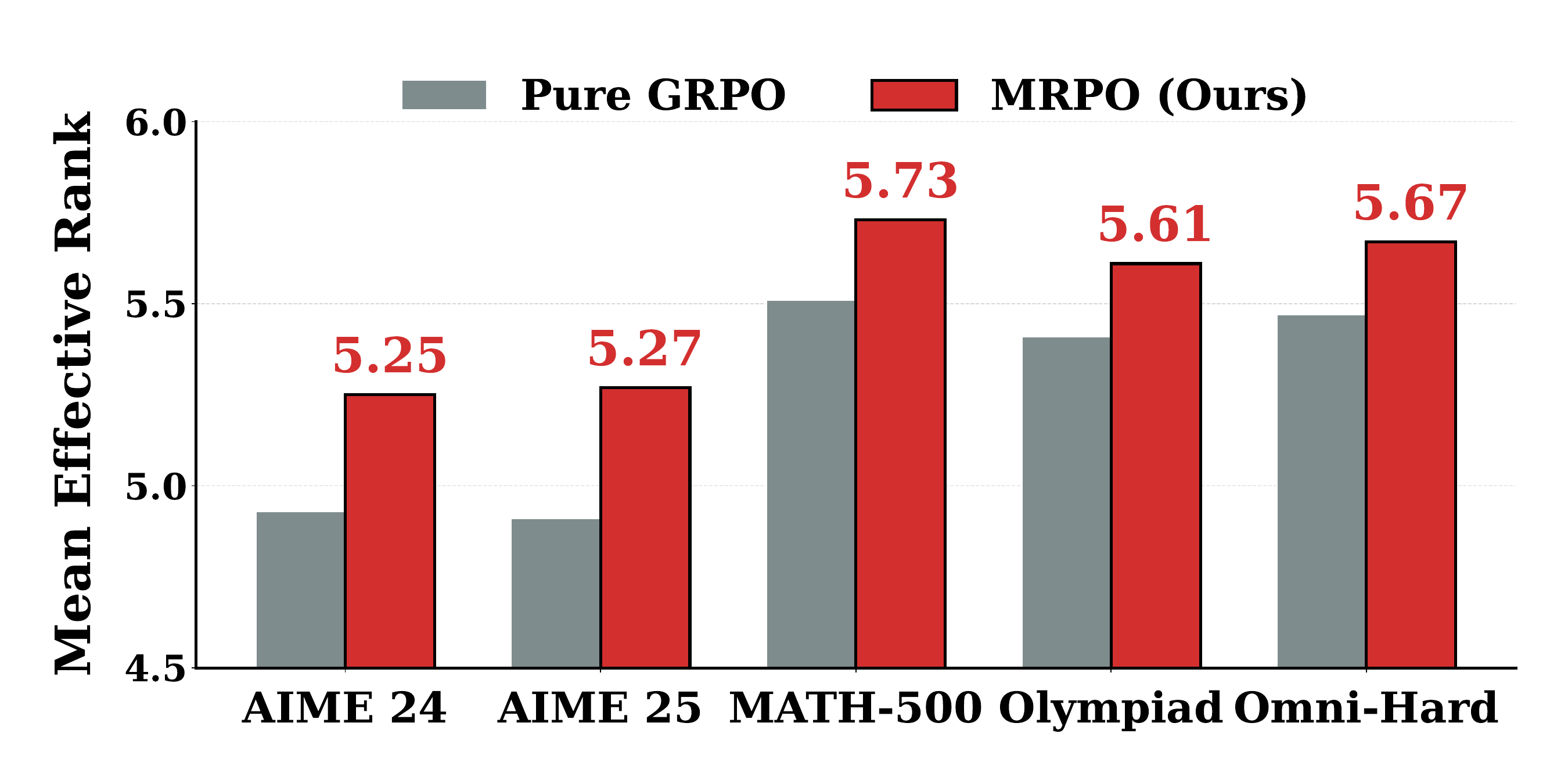}
    \caption{\textbf{Comparison of Rank.}}

        \label{fig:placeholder}
\end{figure}
To empirically validate the efficacy of our Rank-Aware Reward, we calculated the Mean Effective Rank of the reasoning trajectories generated by both Pure GRPO and MRPO across five complex benchmarks. As illustrated in Figure \ref{fig:placeholder}, the baseline Pure GRPO exhibits a marked tendency towards rank collapse, consistently yielding lower-dimensional solution paths particularly on rigorous tasks such as AIME 2024. In contrast, MRPO maintains a significantly higher effective rank across all evaluated domains, achieving a peak spectral dimensionality of 5.73 on MATH-500. These results statistically confirm that our method effectively counteracts the structural degradation typical of standard reinforcement learning, successfully enforcing the preservation of high-rank reasoning geometry throughout the optimization process.

\subsection{Reasoning Boundaries Analysis}

\begin{figure}[h]

    \centering

    \includegraphics[width=0.75\linewidth]{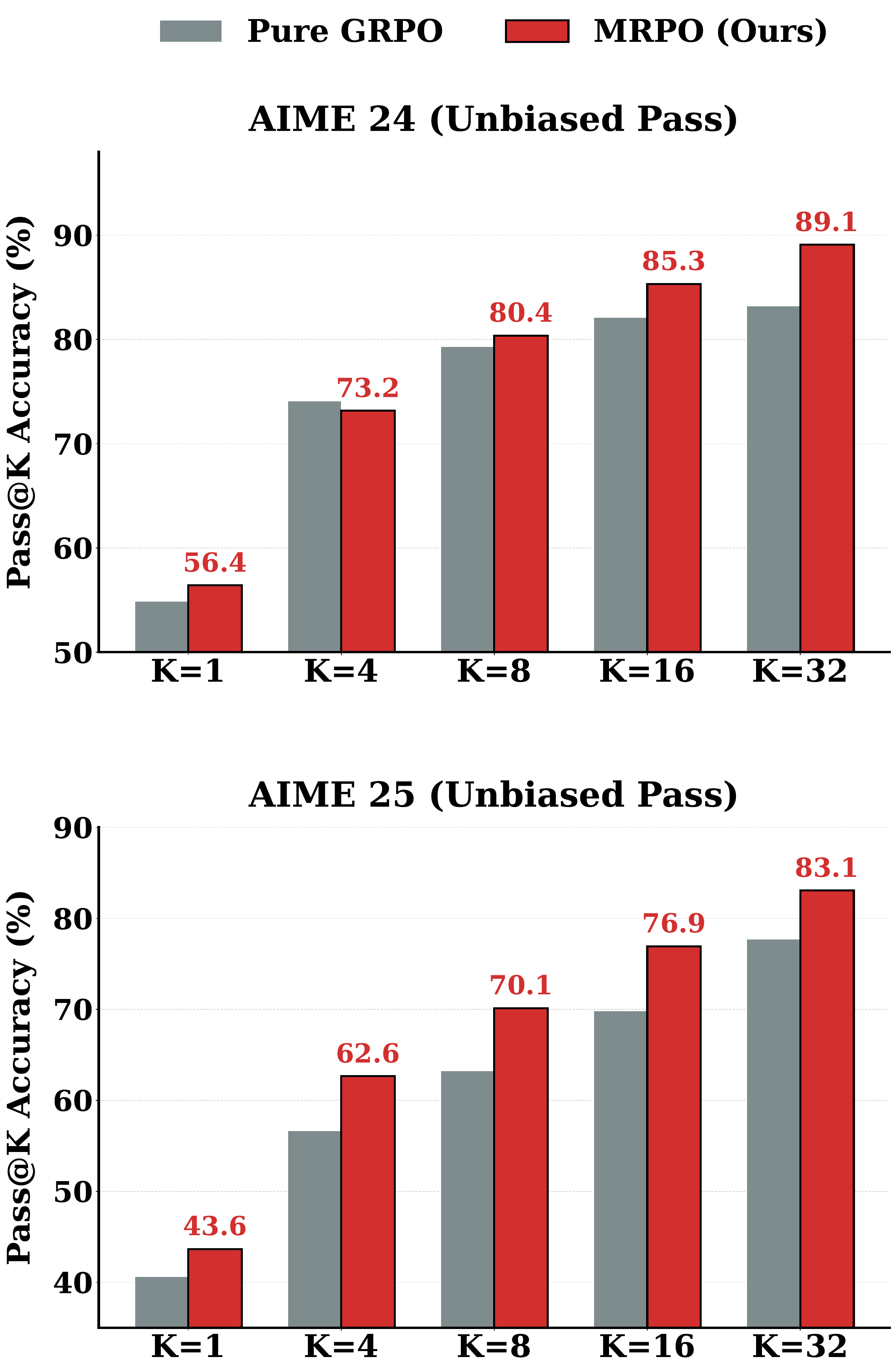}

    \caption{\textbf{Unbiased Pass@K Analysis on AIME Benchmarks.} We evaluate the reasoning coverage of \textbf{MRPO} (Red) versus \textbf{Pure GRPO} (Grey) on AIME 2024 and AIME 2025.}

    \label{fig:aime_pass_k}

\end{figure}

To rigorously evaluate the latent reasoning capacity of our model beyond greedy decoding, we conducted a comprehensive coverage analysis on the AIME 2024 and AIME 2025 benchmarks. For both \textbf{MRPO} and the baseline \textbf{Pure GRPO}, we sampled $n=64$ reasoning trajectories per problem using identical generation parameters (temperature $T=0.7$). Instead of relying on potentially high-variance repeated sampling, we computed the unbiased estimator for Pass@$k$ ($k \in \{1, 4, 8, 16, 32\}$) based on the $n$ generated samples. The unbiased Pass@$k$ is calculated as:

\begin{equation}
    \text{Pass}@k = \mathbb{E}\left[ 1 - \frac{\binom{n-c}{k}}{\binom{n}{k}} \right] = 1 - \frac{\binom{n-c}{k}}{\binom{n}{k}}
\end{equation}

where $n=64$ is the total number of samples and $c$ is the count of correct trajectories.

As visualized in Figure \ref{fig:aime_pass_k}, \textbf{MRPO} demonstrates a statistically significant expansion of the model's capability boundary compared to Pure GRPO. On AIME 2024, our method achieves a remarkable \textbf{Pass@32 of 89.1\%}, indicating that the correct solution lies within the model's accessible search space for nearly 9 out of 10 difficult problems. Similarly, on the more challenging AIME 2025, MRPO maintains a robust \textbf{83.1\%} coverage at $K=32$, consistently outperforming the baseline across all $K$ values.

This consistent superiority suggests that standard RL (GRPO) does not merely suffer from sampling noise, but faces a fundamental "accessibility" limit where correct reasoning paths are geometrically unreachable within the bias manifold. By reshaping this manifold, MRPO effectively lowers the barrier to these correct solutions, converting what were previously "impossible" queries for the baseline into "solvable" ones given sufficient sampling budget.

\subsection{Robustness Analysis}

To verify the reproducibility and stability of our method, we conducted a rigorous robustness analysis by varying the random seeds for both the reinforcement learning training phase and the stochastic sampling during evaluation.

\textbf{Training Stability.} We retrained the MRPO model using two additional random seeds ($2025$ and $2026$) distinct from the primary seed ($42$). As detailed in Table \ref{tab:seed_training} in Appendix B, the model exhibits high stability across all benchmarks. For instance, on the challenging AIME 2024, the performance fluctuates marginally between 53.3\% and 56.7\%, while on MATH-500, the accuracy remains nearly constant (87.0\% - 88.8\%). The consistency of these results confirms that the geometric benefits of MRPO are systematic and not artifacts of lucky initialization.

\textbf{Evaluation Robustness.} We also assessed the sensitivity of our Pass@K estimates on AIME 2024 by conducting independent sampling runs with different seeds. The results, presented in Table \ref{tab:seed_passk} in Appendix B, demonstrate robust coverage estimates. The unbiased Pass@32 metric, for example, remains consistently high (ranging from 86.6\% to 89.1\%), indicating that the expanded "accessibility boundary" discussed in Section 5.3 is a stable property of the reshaped manifold.

\subsection{Efficiency Analysis}
Our empirical analysis confirms that MRPO achieves a highly favorable computational trade-off. It significantly reduces average inference token counts (by approximately 40--60\%) compared to base model while strictly enforcing geometric constraints. Furthermore, the additional overhead from the Rank-Aware Reward calculation is marginal ($<15\%$ of the total iteration time), ensuring minimal impact on wall-clock training efficiency. We provide a detailed breakdown of token consumption and training latency in \textbf{Appendix~\ref{app:efficiency_analysis}}.


\section{Conclusion}
This paper proposes Manifold-Reshaping Policy Optimization (MRPO), a framework that fundamentally challenges prevailing scaling laws by demonstrating that reasoning capacity is governed by \textit{geometric efficiency} rather than parameter count alone. We identify that standard alignment restricts models to a low-rank "Bias Manifold," inevitably leading to reasoning collapse. MRPO overcomes this accessibility boundary by employing Spectral Orthogonal Exploration (SOE) to eject the policy into the "Null Space" and utilizing Effective Rank regularization to sustain high-dimensional reasoning trajectories. Empirically, our 4B parameter model achieves 56.7\% on AIME 2024, significantly surpassing a 32B parameter baseline (33.3\%). These findings validate the existence of a "Geometric Scaling Law," suggesting a paradigm shift in AGI development from parameter expansion to the optimization of latent geometric depth.

\section{Impact Statement}
This paper presents Manifold-Reshaping Policy Optimization (MRPO), a framework that challenges the prevailing scaling laws in Large Language Models (LLMs) by fundamentally redefining the geometric relationship between alignment and reasoning capacity. By leveraging \textbf{Spectral Orthogonal Exploration (SOE)} for cold-start data synthesis and integrating a \textbf{Geometric Reward (Effective Rank)} as a regularization term, we demonstrate—both theoretically and empirically—that reasoning capabilities can be significantly expanded without increasing model size.

\textbf{Computational Efficiency and Environmental Sustainability.} 
Our experimental results indicate that a 4B parameter model trained via MRPO can outperform 32B parameter baselines on complex reasoning benchmarks. This suggests a shift from "parameter scaling" to "geometric scaling," implying that high-level reasoning is achievable with significantly reduced computational resources. Widespread adoption of this paradigm could lower the energy footprint of training and inference, making advanced reasoning agents more sustainable and accessible on consumer-grade hardware.

\textbf{Advancing the Theory of Alignment.} 
We provide a theoretical proof that standard alignment techniques often impose a "geometric ceiling" (Reasoning Collapse) by confining models to a low-rank bias manifold. By proving that "ejecting" the policy into the null space via SOE and stabilizing it with geometric regularization unlocks dormant capacity, we offer a new perspective on the "Alignment Tax." This encourages the research community to view alignment not merely as a constraint satisfaction problem but as a manifold optimization challenge.

\textbf{Safety and Reliability Implications.} 
While our method involves explicitly exploring the "Null Space" outside the pre-trained bias manifold, we argue that this enhances safety in reasoning domains. "Reasoning Collapse" often manifests as hallucination or rote heuristic matching; by enforcing high effective rank, MRPO incentivizes deeper, more robust reasoning chains that are less prone to superficial shortcuts. However, we acknowledge that expanding the inference space requires rigorous oversight. While our Rank-Aware regularization ensures structural coherence, future work must investigate ensuring that the high-dimensional trajectories discovered via SOE remain aligned with human ethical values, particularly when applied to open-ended, non-deterministic domains.

\bibliography{example_paper}
\bibliographystyle{icml2026}
\newpage
\appendix
\onecolumn
\appendix
\section{More Details of Experiment}
\subsection{RL part}
\label{sec:rl_details}

In this section, we provide a comprehensive overview of the experimental setup, data construction, and distributed training infrastructure used for the Reinforcement Learning (RL) fine-tuning stage.

\subsubsection{Algorithm and Framework}
We employ Group Relative Policy Optimization (GRPO) to further align the model's reasoning capabilities with mathematical correctness. Our implementation is based on the OpenRLHF framework. Unlike standard PPO, GRPO eliminates the need for a separate value function (critic) by estimating the baseline from a group of outputs generated from the same prompt. We configure the algorithm with a group size of $G=8$ samples per prompt and utilize group normalization for advantage estimation.

\subsubsection{Training Dataset Construction}
To improve the model's robustness on difficult mathematical problems, we constructed a specialized "hard sample" dataset. We utilized problems from the AIME and AMC competitions (datasets prior to 2023) and the MATH training set.

We performed an initial evaluation using the base model with greedy decoding. We selected approximately 3,000 distinct problems where the model failed to generate the correct answer. This subset represents the frontier of the model's current capability, serving as a high-value training set for the RL stage.

\subsubsection{Distributed Architecture and Engines}
The training is orchestrated using the Ray framework to manage distributed resources efficiently. We utilized a single node equipped with 4 GPUs. The workflow is divided into two primary phases:
\begin{itemize}
    \item \textbf{Inference (Rollout):} We utilize \texttt{vLLM} as the generation engine to accelerate the rollout phase. We configured 4 vLLM engines with a tensor parallel size of 1 and set the GPU memory utilization to 0.6. The communication backend is synchronized via NCCL.
    \item \textbf{Training (Update):} For the policy update phase, we employ DeepSpeed ZeRO-3 (Zero Redundancy Optimizer) to minimize memory redundancy, allowing for the training of the 4B parameter model with BF16 precision.
\end{itemize}

\subsubsection{Hyperparameters and Configuration}
The model was trained for 4 epochs on the selected dataset. We employed a remote reward model based on ranking to provide feedback signals. To stabilize training, the initial KL coefficient was set to 0, relying on the group relative advantage for regularization. The gradient checkpointing and Flash Attention 2 were enabled to optimize memory usage and computational speed.

Table \ref{tab:grpo_params} lists the detailed hyperparameters used in our experiments.

\begin{table}[h]
\centering
\caption{Hyperparameters for GRPO Training via OpenRLHF}
\label{tab:grpo_params}
\begin{tabular}{l|c}
\toprule
\textbf{Hyperparameter} & \textbf{Value} \\
\midrule
\multicolumn{2}{c}{\textit{General Training}} \\
\midrule
Precision & BF16 \\
Optimizer & AdamW (implied) \\
Learning Rate & $1 \times 10^{-6}$ \\
Total Epochs & 4 \\
Global Train Batch Size & 64 \\
Micro Train Batch Size & 2 \\
Gradient Accumulation Steps & Implied by Global/Micro BS \\
Zero Stage & 3 \\
\midrule
\multicolumn{2}{c}{\textit{Generation (Rollout)}} \\
\midrule
Rollout Batch Size & 64 \\
Micro Rollout Batch Size & 8 \\
Prompt Max Length & 4096 \\
Generate Max Length & 4096 \\
Samples per Prompt ($G$) & 8 \\
\midrule
\multicolumn{2}{c}{\textit{GRPO / Algorithm}} \\
\midrule
Advantage Estimator & Group Normalization \\
Init KL Coefficient & 0.0 \\
Reward Normalization & Enabled \\
Attention Implementation & Flash Attention 2 \\
\bottomrule
\end{tabular}
\end{table}

\section{Detailed Efficiency Analysis}
\label{app:efficiency_analysis}

\begin{figure}[H]
    \centering
    \includegraphics[width=0.9\linewidth]{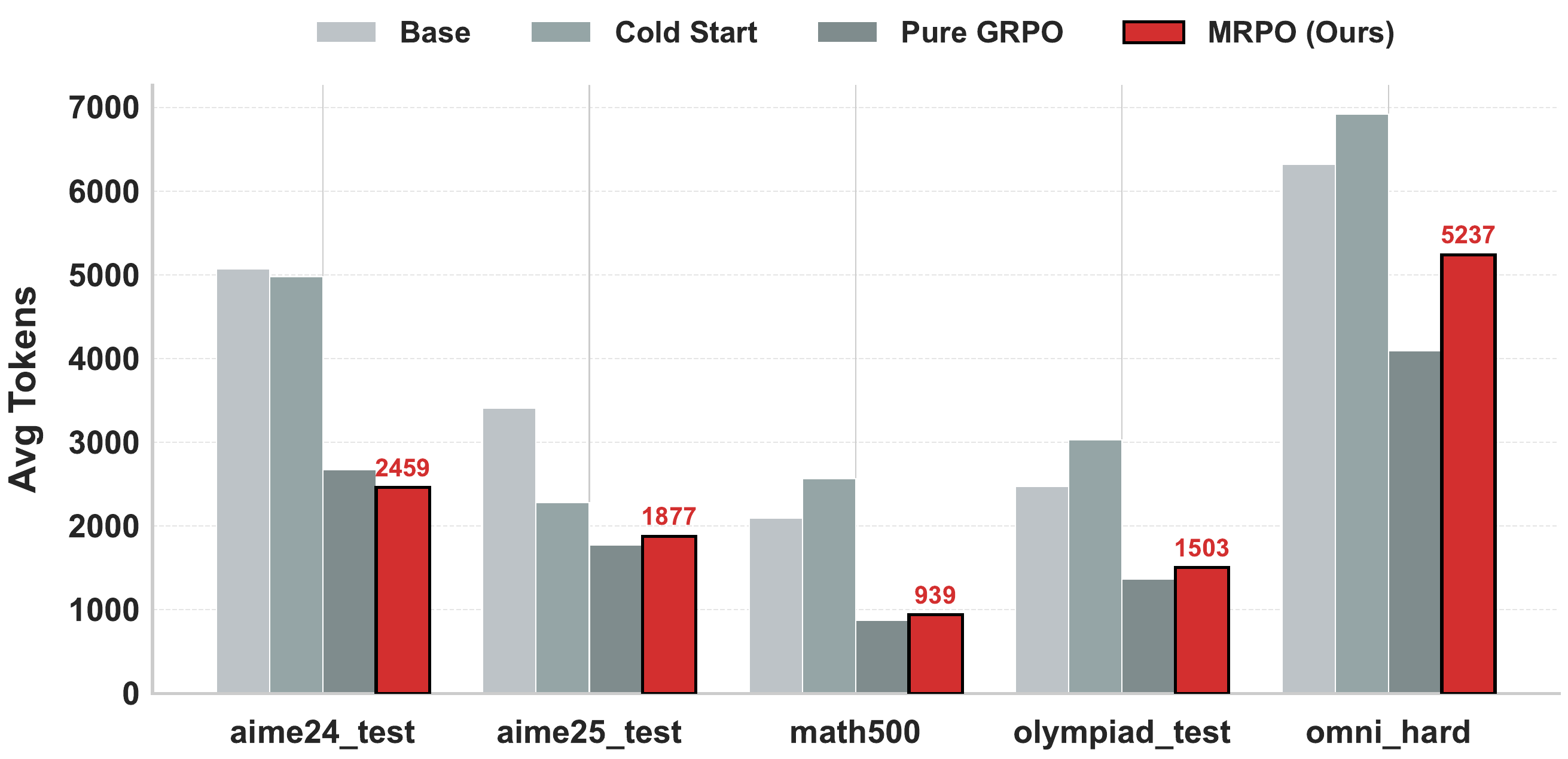}
    \caption{\textbf{Token Efficiency and Preservation of Reasoning Depth.} We analyze the average generation length for problems correctly solved by all models (intersection set).}
    \label{fig:token_analysis}
\end{figure}

In this section, we provide a comprehensive analysis of the computational efficiency of MRPO regarding token consumption and training latency, as visualized in Figure~\ref{fig:token_analysis}.

\textbf{Token Efficiency.} Consistent with the concise optimization nature of reinforcement learning, both MRPO and Pure GRPO significantly reduce average inference token counts by approximately 40--60\% compared to the Base and Cold-Start models. Crucially, while MRPO maintains high-rank reasoning trajectories to prevent collapse, it does not incur a significant ``length tax''; its token consumption remains comparable to Pure GRPO across benchmarks. This indicates that our method enhances geometric depth without reverting to the verbosity of SFT models.

\textbf{Training Latency.} The training loop is dominated by the generation phase. The additional computational overhead introduced by our Rank-Aware Reward calculation is marginal ($<15\%$ of total iteration time) compared to the generation and backpropagation stages. Consequently, MRPO achieves a favorable trade-off, strictly enforcing geometric constraints with minimal impact on wall-clock training efficiency.

\begin{figure}[H]
    \centering
    \includegraphics[width=0.8\linewidth]{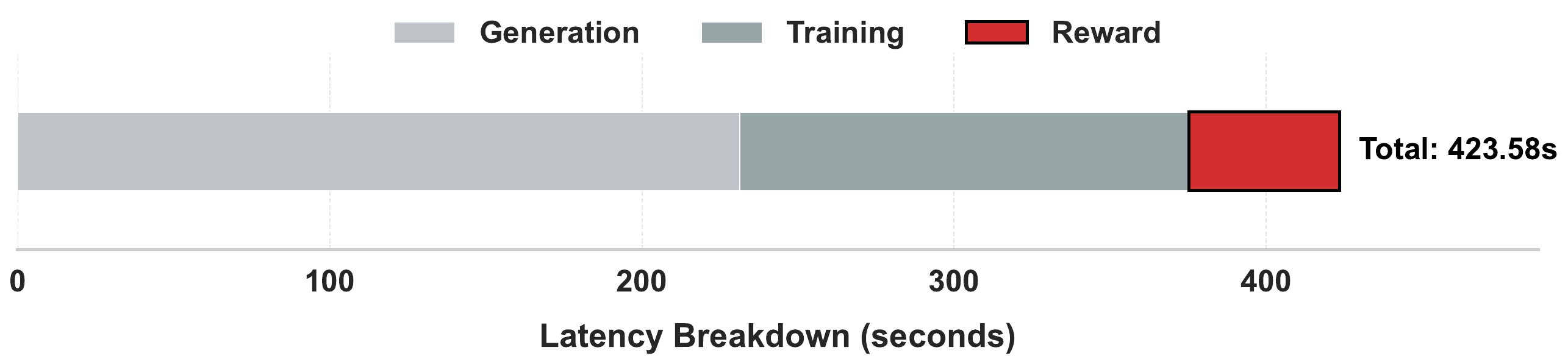}
    \caption{\textbf{System Latency Breakdown.} The time cost analysis (in seconds) per training iteration.}
    \label{fig:latency_breakdown}
\end{figure}

\section{Robustness Experiments}

\begin{table}[H]
\centering
\footnotesize 
\setlength{\tabcolsep}{4pt} 
\caption{\textbf{Robustness of RL Training across Random Seeds.} We compare the Pass@1 accuracy of MRPO trained with three distinct random seeds on five mathematical benchmarks.}
\label{tab:seed_training}
\begin{tabular}{cccccc} 
\toprule
\textbf{Training Seed} & \textbf{AIME 24} & \textbf{AIME 25} & \textbf{MATH-500} & \textbf{OlympiadBench} & \textbf{Omni-Hard} \\
\midrule
42 (Main) & 56.7\% & 43.3\% & 88.8\% & 43.0\% & 17.4\% \\
2025      & 53.3\% & 40.0\% & 87.0\% & 44.2\% & 18.5\% \\
2026      & 53.3\% & 43.3\% & 88.8\% & 43.6\% & 17.8\% \\
\midrule
\bottomrule
\end{tabular}
\end{table}

\begin{table}[H]
\centering
\footnotesize 
\setlength{\tabcolsep}{5pt} 
\caption{\textbf{Robustness of Unbiased Pass@K Estimates on AIME 2024.} We evaluate the variance of accessibility metrics across three different sampling seeds ($n=64$, $T=0.7$).}
\label{tab:seed_passk}
\begin{tabular}{ccccccc} 
\toprule
\textbf{Sampling Seed} & \textbf{K=1} & \textbf{K=4} & \textbf{K=8} & \textbf{K=16} & \textbf{K=32} & \textbf{K=64} \\
\midrule
42   & 56.4\% & 73.2\% & 80.4\% & 85.3\% & 89.1\% & 93.3\% \\
2025 & 56.2\% & 74.3\% & 80.4\% & 84.0\% & 86.6\% & 90.0\% \\
2026 & 55.4\% & 72.6\% & 79.0\% & 83.5\% & 86.6\% & 90.0\% \\
\bottomrule
\end{tabular}
\end{table}

\section{Limitations}
Despite the demonstrated efficacy of MRPO, several limitations persist regarding safety, complexity, and generalization. First, the mechanism of ejecting the policy initialization into the epistemic "Null Space" introduces a trade-off between capability and alignment; by operating outside the pre-trained bias manifold, the model may bypass traditional safety guardrails, potentially increasing the risk of unconstrained or ethical violations. Second, the implementation of the "Student-Guides-Teacher" probe mechanism and real-time manifold estimation via SVD increases the engineering complexity and computational overhead compared to standard RLHF pipelines. Finally, our validation is currently restricted to domains with verifiable rewards (RLVR), and the effectiveness of rank-based geometric regularization in open-ended, subjective, or non-deterministic tasks remains an open question for future research.
\end{document}